\documentclass[conference]{IEEEtran}
\IEEEoverridecommandlockouts

\usepackage{cite}
\usepackage{amsmath,amssymb}
\usepackage{graphicx}
\usepackage{url}
\usepackage{tikz}
\usetikzlibrary{positioning,fit,backgrounds,arrows.meta,calc}
\usepackage{hyperref}

\begin{document}

\title{Towards an Intention Abstraction Layer for\\ Autonomous Industrial Systems}

\author{
  
\IEEEauthorblockN{Artan Markaj\thanks{\copyright~2026 IEEE. Personal use of this material is permitted. Permission from IEEE must be obtained for all other uses, in any current or future media, including reprinting/republishing this material for advertising or promotional purposes, creating new collective works, for resale or redistribution to servers or lists, or reuse of any copyrighted component of this work in other works.}}
\IEEEauthorblockA{\textit{Eurogate GmbH \& Co. KGaA, KG} \\
Hamburg, Germany\\
artan.markaj@eurogate.eu}

\and

\IEEEauthorblockN{Raphael Höfer}
\IEEEauthorblockA{\textit{Helmut Schmidt University Hamburg} \\
Hamburg, Germany\\
raphael.hoefer@hsu-hh.de}

\and

\IEEEauthorblockN{Felix Gehlhoff}
\IEEEauthorblockA{\textit{Helmut Schmidt University Hamburg} \\
Hamburg, Germany\\
felix.gehlhoff@hsu-hh.de}
}

\maketitle

\begin{abstract}
Modern industrial environments increasingly run many autonomous subsystems
at once -- schedulers, energy managers, vehicle fleets -- each pursuing its own
goals while sharing the same physical resources. Because high-level human
intentions are translated into low-level control logic and then discarded, no
running component can tell whether it is still doing what was actually
intended, and goal conflicts surface only after they have caused a missed
target or a shutdown. We propose the \emph{Intention Abstraction Layer} (IAL),
a domain-agnostic middleware that represents intentions as first-class,
persistent, and explainable runtime objects: a large language model grounded
in a formal OWL ontology parses natural-language goals into structured
intentions, a consistency monitor detects conflicts at registration
time, before execution, and a transparency module explains them in natural
language. We report a first proof of concept in which two autonomous agents
register conflicting production and energy intentions, and the IAL flags and
explains the conflict before it reaches the execution layer. The result is a
mechanism that shifts behavioral assurance for cooperating autonomous systems
from post-hoc failure analysis to pre-execution, intention-level checking.
\end{abstract}


\section{Introduction}
Industrial automation is shifting from monolithic control toward
environments populated by many autonomous subsystems that operate
concurrently -- production schedulers, energy-management agents, automated
guided vehicle (AGV) fleets, and maintenance planners -- each optimizing its
own objective while competing for the same physical resources such as
energy, floor space, equipment, and time. As autonomy increases, the
behavior of such a system is no longer the product of a single controller
but the emergent result of many independently specified goals. Building on a
prior intention-based engineering framework for modular process
plants~\cite{markaj2022intention} and an autonomous-systems environment for
evaluating control in autonomous plant operation~\cite{markaj2024aste}, this
paper lifts intention modeling from the design phase into runtime operation
and generalizes it beyond process plants to autonomous industrial systems at
large.

A structural problem accompanies this shift. Goals are formulated by humans
at a high level of abstraction (``maximize throughput for Order~A while
keeping energy below 80\,\% of peak load''), but they are executed at the
low level of setpoints and control logic. In this translation the
rationale, the \emph{why} behind a goal, is discarded, so no component of a
running system can answer whether it is still pursuing what was actually
intended. Three failure patterns recur. \emph{Lost intent}: the original
objective survives only as setpoints, and its purpose can no longer be
queried at runtime. \emph{Silent conflict}: two agents specify objectives
independently, neither aware of the other, and the contradiction surfaces
only once a physical constraint is violated or a target is missed.
\emph{Stale intent}: an objective set at $t_0$ keeps executing at $t_3$ even
though its context has changed (e.g., the order was cancelled, the product
discontinued) so the system acts consistently against an obsolete goal.
Today these reconciliations are still performed implicitly: integration
engineers resolve conflicting objectives at design time, and supervising
operators catch drift by watching the process. The problem we address is
therefore an \emph{emerging} one, as autonomy removes the human from the
loop, that implicit safeguard disappears and is given no formal home.

Existing coordination mechanisms do not close this gap because none
represents intentions as managed runtime objects: contract-net and auction
protocols are transactional and carry no notion of \emph{why} a resource is
requested~\cite{smith1980contractnet}; agent communication languages such as
FIPA-ACL exchange goal-bearing messages but neither persist nor cross-check
them at runtime~\cite{fipa2002acl}; and rule engines detect only conflicts
anticipated at design time. What is missing, we argue, is a runtime
infrastructure that treats intentions as first-class, persistent, and
explainable resources across agent boundaries. At the same time, large
language models have begun to recognize and discover intent directly from
natural language~\cite{rodriguez2024intentgpt}, pointing to a practical route
for capturing intentions at runtime.

This work-in-progress paper proposes such an infrastructure -- the
\emph{Intention Abstraction Layer} (IAL) -- and reports a first proof of
concept. Our contributions are: (i)~the IAL architecture, a domain-agnostic
middleware that treats intentions as first-class runtime resources and
manages them along two axes: \emph{elicitation} of intentions between human
operators and autonomous agents, and \emph{reconciliation} of intentions
among the agents themselves, with formal capture, persistence, and lifecycle
management; (ii)~LLM-grounded intention parsing and conflict
detection, showing that a language model anchored to a formal OWL ontology
can bridge natural-language goal specification and machine-interpretable
intention structures and can flag goal conflicts in open-world scenarios
without pre-specified conflict rules; and (iii)~an open-source proof-of-concept
that validates the three core modules on a multi-agent production scenario.

\section{Background and Related Work}
Our work connects three strands of research.

\textit{Goal- and intention-oriented modeling.} Goal-oriented requirements
engineering captures stakeholder objectives and their refinement, e.g., the
KAOS method~\cite{vanlamsweerde2001goal}, while the belief--desire--intention
(BDI) model~\cite{rao1995bdi} gives agents an explicit notion of goals and is
still being extended, for instance by integrating machine
learning~\cite{agiollo2025bdi}. These approaches formalize intentions at
design time or inside individual agents; we instead manage them as shared,
living runtime objects that can be registered, monitored, and retired across
agents. Our
own prior work introduced ontology-based intention modeling for the early
engineering of modular process plants~\cite{markaj2022intention} and methods
for evaluating the behavior of autonomous plants~\cite{markaj2024operator};
here we lift intention modeling into runtime operation.

\textit{Coordination in multi-agent systems.} Classical mechanisms coordinate
behavior without a shared representation of \emph{why}: the Contract Net
protocol allocates tasks through bidding~\cite{smith1980contractnet}, and
agent communication languages such as FIPA-ACL exchange goal-bearing
messages~\cite{fipa2002acl}, yet none persists, monitors, or cross-checks
intentions across agents at runtime. Recent benchmarks confirm the
difficulty: even LLM-based agents coordinate well on observable conditions
but struggle to reason about their partners' beliefs and
intentions~\cite{agashe2025llmcoord}. Closing this gap at runtime is precisely
what the IAL targets.

\textit{Language models in industrial automation.} A rapidly growing body of
work applies LLMs to production and control, enhancing modular
production~\cite{xia2023flexible}, controlling automation systems
directly~\cite{xia2025control}, exploring resource capabilities in multi-agent
manufacturing~\cite{lim2025resource}, building agentic
frameworks~\cite{vyas2025autonomous}, and integrating LLMs with digital
twins~\cite{xia2025architecture}. Closest to our parser, NL2IBE maps natural
language to formalized engineering artefacts under ontology
control~\cite{schoch2024nl2ibe}; we apply the same principle to runtime
\emph{intentions} rather than design-time artefacts. The capability-and-skill
paradigm, increasingly paired with LLMs~\cite{vieiradasilva2025mcp},
formalizes what a system \emph{can do}; the IAL is complementary, addressing
what it \emph{intends}. We ground the model in an OWL
ontology~\cite{mcguinness2004owl} and keep it strictly \emph{above} execution.

\section{Problem Statement and Research Questions}
The failure patterns above share a common root: intentions exist only
implicitly, dispersed across operator knowledge, configuration files, and
control code, and are never reconciled while the system runs -- neither
between the human operators who set goals and the agents that execute them,
nor among the agents themselves. No shared, formal, runtime representation of
\emph{what each party currently intends} exists, so divergence between
intended and executed behavior becomes observable only through its
consequences. We therefore ask:

\begin{itemize}
  \item[\textbf{RQ1}] Can a dedicated middleware layer -- the IAL -- formally
  capture, persist, and monitor the intentions exchanged between human
  operators and autonomous agents and among the agents themselves, and detect
  conflicts \emph{before} they manifest as execution failures?
  \item[\textbf{RQ2}] Can LLM-based parsing bridge natural-language goal
  specifications and formal, ontology-based intention representations well
  enough to make the IAL practically usable in industrial settings?
\end{itemize}

\noindent This paper treats the \emph{forward} direction, from stated goals
to monitored, reconciled intentions. We identify the \emph{backward}
direction, inferring the intention implied by a planned execution and
checking declared against enacted behavior, as the bridge to our subsequent
work, and sketch it as a capability of the architecture below.

\section{Proposed Approach}
\begin{figure}[t]
  \centering
\resizebox{\columnwidth}{!}{%
\begin{tikzpicture}[
  font=\footnotesize,
  layer/.style={draw, rounded corners, minimum width=8.2cm,
                minimum height=0.95cm, align=center, fill=black!4},
  mod/.style={draw, rounded corners, minimum width=2.35cm,
              minimum height=1.0cm, align=center, fill=white, inner sep=2pt},
  sub/.style={draw, rounded corners, minimum width=7.5cm,
              minimum height=0.6cm, align=center, fill=black!7},
  fwd/.style={-{Latex[length=2mm]}, semithick},
  bwd/.style={-{Latex[length=2mm]}, semithick, densely dashed},
  lbl/.style={font=\scriptsize, inner sep=1.5pt},
]
  \node[layer] (top) {\textbf{Human Operator / Agent Layer}\\[-2pt]
                      \scriptsize natural-language or structured goal input};

  \node[mod, below=1.9cm of top] (mon) {Consistency\\Monitor};
  \node[mod, left=0.18cm of mon]  (parser) {Intent Parser\\\scriptsize LLM+Ontology};
  \node[mod, right=0.18cm of mon] (api) {Transparency\\\& Resolution};
  \node[font=\footnotesize\bfseries, above=1mm of mon] (ialtitle)
        {Intention Abstraction Layer (IAL)};
  \node[sub, below=0.25cm of mon] (store)
        {Intention Store $+$ Ontology/Reasoner (OWL)};

  \begin{scope}[on background layer]
    \node[draw, rounded corners, fill=blue!5, inner sep=0.28cm,
          fit=(parser)(api)(store)(ialtitle)]
          (ial) {};
  \end{scope}

  \node[layer, below=1.5cm of ial] (exec) {\textbf{Execution Layer}\\[-2pt]
        \scriptsize PLC / agent framework / vehicle controller
        (capabilities \& skills)};

  \draw[fwd] ($(top.south)+(-1.4,0)$) -- ($(ial.north)+(-1.4,0)$)
        node[lbl, midway, left, align=right] {elicit\\(NL$\to$intention)};
  \draw[bwd] ($(ial.north)+(1.4,0)$) -- ($(top.south)+(1.4,0)$)
        node[lbl, midway, right, align=left] {explain /\\confirm};

  \draw[fwd] ($(ial.south)+(-1.4,0)$) -- ($(exec.north)+(-1.4,0)$)
        node[lbl, midway, left, align=right] {register\\(OPC-UA/MQTT)};
  \draw[bwd] ($(exec.north)+(1.4,0)$) -- ($(ial.south)+(1.4,0)$)
        node[lbl, midway, right, align=left] {infer from\\execution};
\end{tikzpicture}%
}
  \caption{The IAL sits between goal specification and execution.
  Solid arrows (\emph{forward} path): intentions are elicited (NL$\to$ontology),
  reconciled, and registered toward execution. Dashed arrows
  (\emph{backward} path): decisions explained to operators and intentions
  inferred from execution; sketched here, developed in subsequent work.}
  \label{fig:concept}
\end{figure}
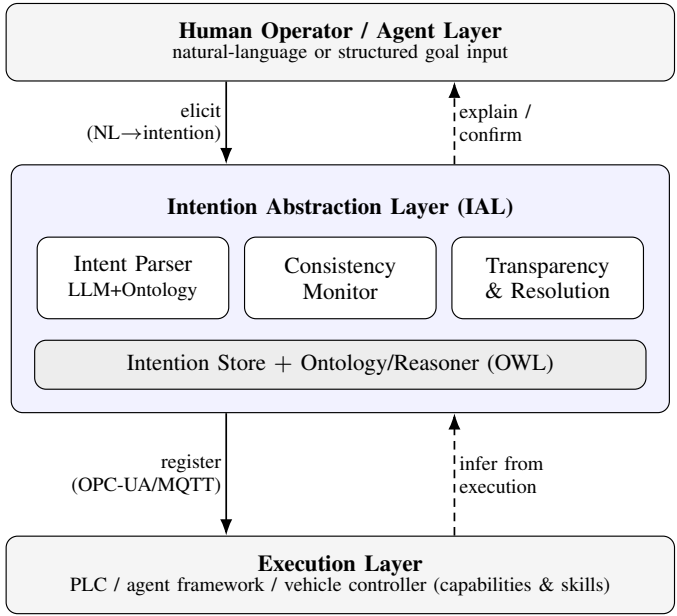

The IAL is a domain-agnostic middleware layer positioned between the
goal-specification level (human operators or autonomous agents) and the
execution level (PLCs, controllers, agent execution frameworks), as shown in
Fig.~\ref{fig:concept}. It is deliberately a coordination and transparency
layer rather than a safety-critical controller: it reasons about and
reconciles intentions \emph{above} the execution layer and never intervenes
in real-time control directly. This boundary keeps non-deterministic
components -- notably the language model -- out of the safety-critical path.

\subsection{Formal intention representation}
In the underlying intention-based engineering
framework~\cite{markaj2022intention}, an intention is described by
\emph{intentional elements}: \emph{Goals} (what is to be achieved),
\emph{Abstract Solutions}, strategies, stated at a solution-neutral
level, and \emph{Requirements}, the constraints that must hold. For runtime
operation we add \emph{Context}, the operating situation in which the
intention is valid, together with provenance metadata, giving
\begin{equation}
  \mathit{Intention} = \langle\, \underbrace{G}_{\text{Goal}},\;
  \underbrace{A}_{\text{Abstr.\,Sol.}},\;
  \underbrace{R}_{\text{Requirement}},\;
  \underbrace{C}_{\text{Context}} \rangle\;
  \label{eq:intention}
\end{equation}
where $G$, $A$ and $R$ are ontology classes taken from the framework and the
context $C$ is introduced here. Because each element is an ontology object
rather than free text, an intention is machine-interpretable and reasoning
over intentions reduces to reasoning over the ontology.

\subsection{Intent Parser (LLM + ontology)}
The Intent Parser maps natural-language or semi-structured input to an
instance of~\eqref{eq:intention}. A large language model performs the
mapping, but it is \emph{anchored} to the ontology: the target vocabulary for
goals, abstract solutions, and requirements is constrained to ontology terms,
so the parser produces grounded structures rather than unconstrained text.
This is
what answers RQ2 -- it lets operators and agents state goals in their own
terms while the IAL still obtains a formal representation.

\subsection{Consistency Monitor}
On registration, the Consistency Monitor checks the new intention against all
currently active intentions in the store using ontology-based semantic
similarity and constraint intersection. When two intentions act on
overlapping resources and time windows with incompatible constraints, it
raises a conflict, annotated with type, severity, and the affected agents.
Crucially, this happens at \emph{registration} time, before execution, which
is the mechanism behind RQ1 and the pre-execution behavioral-evaluation angle.

\subsection{Transparency API and resolution}
The Transparency API renders detected conflicts and IAL decisions back into
operator-facing natural language, closing the loop opened by the parser. It
turns formal ontology reasoning into an explanation a human can act on and,
where one exists, proposes a candidate resolution, for example, delaying a
batch or relaxing a temperature ramp. Consistent with the design boundary,
the IAL \emph{advises} but does not enforce: a human or the owning agent
accepts or rejects the proposal.

\subsection{Intention Store and lifecycle}
The Intention Store is a persistent OWL repository of all active and
historical intentions across agents, backed by the ontology and a reasoner
that the Consistency Monitor queries. Because intentions are explicit objects,
they have a lifecycle: they can be updated, superseded, or retired. This
directly addresses \emph{stale intent}, the failure pattern that arises when
no one manages the validity of an objective over time.

\subsection{Directionality}
The IAL operates along two axes. In the \emph{forward} direction it elicits
and reconciles intentions: an operator or agent states a goal, the Parser
lifts it into~\eqref{eq:intention}, and the Monitor reconciles it against the
active set---the path exercised in this paper. In the \emph{backward}
direction the layer inverts the mapping, inferring the intention implied by an
observed execution and comparing it against the declared intention to expose
divergence between what an agent says and what it does; we sketch this
direction here and develop it in subsequent work.

\section{Preliminary Implementation and First Results}
We implemented a minimal proof of concept whose modules map one-to-one onto
Fig.~\ref{fig:concept}: an Intent Parser, a Consistency Monitor, a
Transparency and Resolution module, and an OWL-backed Intention Store, sitting
on top of a \emph{simulated execution layer}. The ontology is an OWL file
built with \texttt{owlready2}. Consistent with the design boundary, the
language model is confined to the parser; all conflict logic is deterministic
Python.

The simulated execution layer is a minimal process model of the plant's
aggregate load on a shared electrical bus: other consumers follow a
time-varying baseline profile, and an executing high-temperature batch
superimposes its demand on top, producing a load trajectory that the IAL
evaluates against the active constraint \emph{before} execution. It is
deliberately simple -- a software stand-in for a PLC or control system, not a
high-fidelity plant model.

\subsection{Scenario}
Two autonomous agents register intentions concurrently. The
\emph{Production Scheduler} states: ``run a high-temperature batch from
14:00--16:00 that draws about 92\,\% of peak load''; the \emph{Energy Manager}
states: ``keep total load below 80\,\% of peak between 13:00--17:00.'' Each
sentence is parsed into the structure of~\eqref{eq:intention} by Anthropic's
Claude (Opus~4.8), constrained to the ontology vocabulary so that its output
is grounded rather than free text; the parsed structures are recorded so that
the reported result is deterministic and reproducible.

\subsection{Result}
\begin{figure}[t]
  \centering
  \includegraphics[width=\columnwidth]{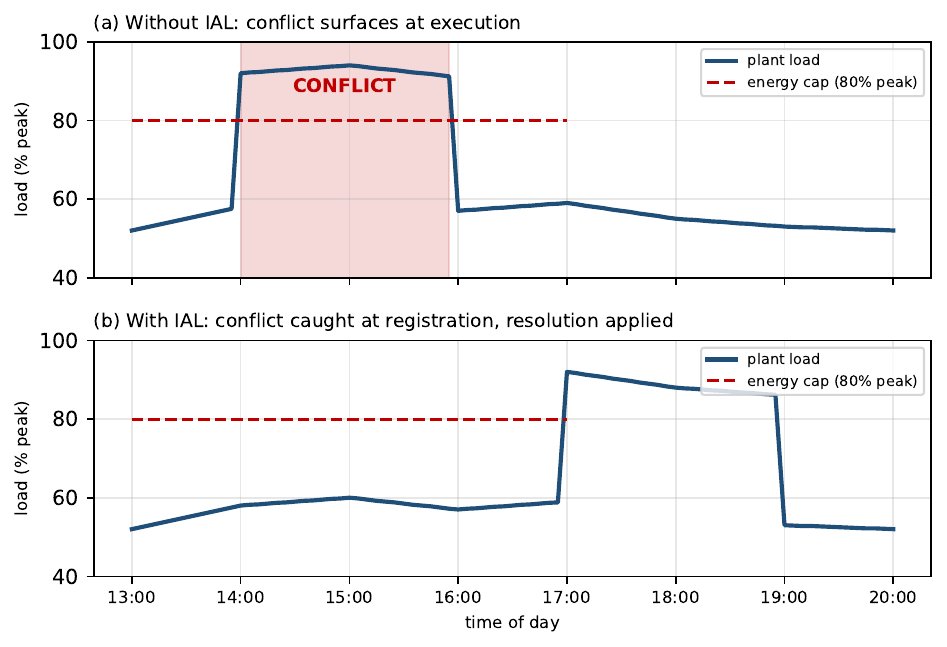}
  \caption{Simulated plant load for the production/energy scenario.
  (a)~Without the IAL, the batch executes and breaches the 80\,\% cap during
  14:00--16:00. (b)~With the IAL, the conflict is detected at intention
  registration and the proposed resolution defers the batch past the cap
  window, keeping the load compliant. All values are produced by running the
  code, not hand-set.}
  \label{fig:result}
\end{figure}

On registering the second intention, the Consistency Monitor reports a
\emph{high}-severity conflict on the shared resource \texttt{peak\_load}: the
batch demands $\sim$92\,\% during 14:00--16:00, exceeding the active
80\,\% cap by 12 percentage points in the overlapping window, the exceedance
margin by which the monitor grades severity. The Transparency
module explains this in natural language and proposes a resolution: defer the
batch until the cap window ends (17:00) or reduce its demand. Evaluating both
cases against the simulated execution layer yields Fig.~\ref{fig:result}:
\emph{without} the IAL the conflict is latent and the cap is breached at
execution (a); \emph{with} the IAL it is caught at registration and the
resolved schedule keeps the load under the cap throughout the constrained
window (b). The conflict is thus surfaced and explained \emph{before}
execution rather than after a violation, the behavior targeted by RQ1, with
the parsing step answering RQ2.

\subsection{Application}
To apply the IAL, an automation engineer deploys it as a coordination service
\emph{alongside} an existing control system or agent framework.
The ontology is extended once with the plant's vocabulary; the system is then
connected in one of two ways. In the \emph{active} mode each autonomous
component calls \texttt{register(\textit{goal},\,\textit{source})} when it
forms a plan and subscribes to conflict notifications over OPC-UA, REST, or
MQTT. In the \emph{passive} mode the IAL observes setpoints and schedules and
infers intentions without modifying the components. Detected conflicts and
their explanations are routed to the operator's HMI or returned to the agents
as advisory signals; the control system keeps running throughout, as the IAL
only observes, reconciles, and explains.

The implementation will be released as open source as a next step.

\section{Conclusion and Future Work}
We presented the Intention Abstraction Layer, a middleware that captures,
reconciles, and explains the intentions of multiple autonomous systems at
runtime, together with a proof of concept that detects and explains a
production/energy conflict before execution. The work is preliminary. The
language model is non-deterministic and adds latency, so we confine it to the
advisory layer and record its output; the proof of concept implements only the
forward path over a simulated execution layer. Real PLC/OPC-UA connectivity,
agent identity and trust, and concurrency control remain open engineering
challenges. A rule-based or locally-deployable model alternative to the cloud
LLM is planned to address determinism and on-premise deployment requirements.
A confirmation step for the operator even in the conflict-free case, verifying
that the parsed intention matches the stated goal before execution, is
identified as a trust-building measure for deployment. The \emph{backward} direction -- inferring intentions from observed
execution to check declared against enacted behavior -- remains an open
direction for future work.

\bibliographystyle{IEEEtran}
\bibliography{bibliography}

\begin{thebibliography}{10}
\providecommand{\url}[1]{#1}
\csname url@samestyle\endcsname
\providecommand{\newblock}{\relax}
\providecommand{\bibinfo}[2]{#2}
\providecommand{\BIBentrySTDinterwordspacing}{\spaceskip=0pt\relax}
\providecommand{\BIBentryALTinterwordstretchfactor}{4}
\providecommand{\BIBentryALTinterwordspacing}{\spaceskip=\fontdimen2\font plus
\BIBentryALTinterwordstretchfactor\fontdimen3\font minus
  \fontdimen4\font\relax}
\providecommand{\BIBforeignlanguage}[2]{{%
\expandafter\ifx\csname l@#1\endcsname\relax
\typeout{** WARNING: IEEEtran.bst: No hyphenation pattern has been}%
\typeout{** loaded for the language `#1'. Using the pattern for}%
\typeout{** the default language instead.}%
\else
\language=\csname l@#1\endcsname
\fi
#2}}
\providecommand{\BIBdecl}{\relax}
\BIBdecl

\bibitem{markaj2022intention}
A.~Markaj and A.~Fay, ``Intention-based engineering for the early design phases
  and the automation of modular process plants,'' in \emph{2022 IEEE 27th
  International Conference on Emerging Technologies and Factory Automation
  (ETFA)}, Stuttgart, Germany, Sep. 2022.

\bibitem{markaj2024aste}
A.~Markaj, M.~Mercang{\"o}z, and A.~Fay, ``Design and implementation of an
  autonomous systems training environment framework for control algorithm
  evaluation in autonomous plant operation,'' \emph{Computers \& Chemical
  Engineering}, vol. 189, p. 108798, Oct. 2024.

\bibitem{smith1980contractnet}
R.~G. Smith, ``The contract net protocol: High-level communication and control
  in a distributed problem solver,'' \emph{IEEE Transactions on Computers},
  vol. C-29, no.~12, pp. 1104--1113, 1980.

\bibitem{fipa2002acl}
{Foundation for Intelligent Physical Agents}, ``{FIPA} {ACL} message structure
  specification,'' FIPA, Tech. Rep. SC00061G, 2002.

\bibitem{rodriguez2024intentgpt}
J.~A. Rodriguez, N.~Botzer, D.~Vazquez, C.~Pal, M.~Pedersoli, and I.~Laradji,
  ``{IntentGPT}: Few-shot intent discovery with large language models,'' in
  \emph{ICLR 2024 Workshop on Large Language Model (LLM) Agents}, 2024.

\bibitem{vanlamsweerde2001goal}
A.~van Lamsweerde, ``Goal-oriented requirements engineering: A guided tour,''
  in \emph{Proceedings Fifth IEEE International Symposium on Requirements
  Engineering (RE'01)}, 2001, pp. 249--262.

\bibitem{rao1995bdi}
A.~S. Rao and M.~P. Georgeff, ``{BDI} agents: From theory to practice,'' in
  \emph{Proceedings of the First International Conference on Multi-Agent
  Systems (ICMAS-95)}, 1995, pp. 312--319.

\bibitem{agiollo2025bdi}
A.~Agiollo and A.~Omicini, ``Integrating machine learning into
  belief-desire-intention agents: Current advances and open challenges,''
  \emph{arXiv preprint arXiv:2510.20641}, 2025.

\bibitem{markaj2024operator}
A.~Markaj, F.~Pelzer, N.~Richter, M.~Mercang{\"o}z, and A.~Fay, ``Development
  and integration of operator behavior models for the evaluation of autonomous
  plants,'' in \emph{2024 IEEE 29th International Conference on Emerging
  Technologies and Factory Automation (ETFA)}, Padova, Italy, 2024.

\bibitem{agashe2025llmcoord}
S.~Agashe, Y.~Fan, A.~Reyna, and X.~E. Wang, ``{LLM}-coordination: Evaluating
  and analyzing multi-agent coordination abilities in large language models,''
  in \emph{Findings of the Association for Computational Linguistics: NAACL
  2025}, 2025.

\bibitem{xia2023flexible}
Y.~Xia, M.~Shenoy, N.~Jazdi, and M.~Weyrich, ``Towards autonomous system:
  Flexible modular production system enhanced with large language model
  agents,'' in \emph{2023 IEEE 28th International Conference on Emerging
  Technologies and Factory Automation (ETFA)}, 2023.

\bibitem{xia2025control}
Y.~Xia, N.~Jazdi, J.~Zhang, C.~Shah, and M.~Weyrich, ``Control industrial
  automation systems with large language models,'' in \emph{2025 IEEE 30th
  International Conference on Emerging Technologies and Factory Automation
  (ETFA)}, 2025.

\bibitem{lim2025resource}
J.~Lim and I.~Kovalenko, ``A large language model-enabled control architecture
  for dynamic resource capability exploration in multi-agent manufacturing
  systems,'' \emph{arXiv preprint arXiv:2505.22814}, 2025.

\bibitem{vyas2025autonomous}
J.~Vyas and M.~Mercang{\"o}z, ``Autonomous control leveraging {LLMs}: An
  agentic framework for next-generation industrial automation,'' \emph{arXiv
  preprint arXiv:2507.07115}, 2025.

\bibitem{xia2025architecture}
Y.~Xia, N.~Jazdi, and M.~Weyrich, ``An architecture for integrating large
  language models with digital twins and automation systems,'' in \emph{2025
  IEEE 30th International Conference on Emerging Technologies and Factory
  Automation (ETFA)}, 2025.

\bibitem{schoch2024nl2ibe}
N.~Schoch and M.~Hoernicke, ``{NL2IBE}: Ontology-controlled transformation of
  natural language into formalized engineering artefacts,'' in \emph{2024 IEEE
  Conference on Artificial Intelligence (CAI)}, 2024, pp. 997--1004.

\bibitem{vieiradasilva2025mcp}
L.~M. Vieira~da Silva, A.~K{\"o}cher, and F.~Gehlhoff, ``Beyond formal
  semantics for capabilities and skills: Model context protocol in
  manufacturing,'' in \emph{2025 IEEE 30th International Conference on Emerging
  Technologies and Factory Automation (ETFA)}, 2025.

\bibitem{mcguinness2004owl}
D.~L. McGuinness and F.~van Harmelen, ``{OWL} web ontology language overview,''
  W3C, W3C Recommendation, 2004.

\end{thebibliography}

\end{document}